\documentclass[journal]{IEEEtran}
\usepackage{graphicx}
\usepackage{amsmath,amssymb}
\usepackage{booktabs}
\usepackage{makecell,multirow,array}
\usepackage{comment}
\usepackage{xcolor}
\usepackage{cite}
\usepackage[caption=false,font=footnotesize]{subfig}
\usepackage{xspace}

\usepackage{placeins}
\usepackage{stfloats}
\usepackage[hidelinks]{hyperref}
\usepackage[capitalise,noabbrev]{cleveref}
\title{SL-Ground: Sentence-Level Grounding for Fine-Grained Open-Vocabulary Object Detection}
\author{Yang-Syu Li, Cheng-Huan Yang, Chia-Yung Jui, and Peggy Joy Lu%
\thanks{Yang-Syu Li, Cheng-Huan Yang and Peggy Joy Lu are with the Department of Computer Science and Information Engineering, National Chung Cheng University, Chiayi, Taiwan. }%
\thanks{Chia-Yung Jui is with the National Center for High-Performance Computing, Taichung, Taiwan.}
\thanks{Peggy Joy Lu is the corresponding author (e-mail: peggylu@cs.ccu.edu.tw).}
}

\begin{document}
\maketitle
\begin{abstract}
Grounding-based open-vocabulary detectors provide a widely used framework for text-conditioned localization, with a phrase-grounding formulation that aligns object queries to individual text tokens and recovers class scores through token-to-class aggregation. We observe that this token-level scoring design becomes a bottleneck for fine-grained open-vocabulary object detection, where the target category is defined by a complete compositional description rather than by isolated tokens. Based on this observation, we propose SL-Ground, a sentence-level grounding reformulation for grounding-based open-vocabulary detection. Instead of scoring queries against fragmented token evidence, SL-Ground encodes each category description as a complete sentence-level representation and directly scores object queries against these representations. To evaluate fine-grained recognition across label granularity, we construct a multi-granularity Stanford Cars benchmark spanning vehicle type, brand, and car model. At the finest car-model granularity, SL-Ground raises AP over the token-level variant from 14.5 to 70.6 under matched training candidate sampling and to 86.0 with full-vocabulary training, while better preserving open-vocabulary capability and avoiding repeated prompt chunking. Beyond Stanford Cars, SL-Ground also improves FG-OVD performance, achieving the best average AP among compared methods. These results show that sentence-level grounding provides a more suitable scoring unit for fine-grained OVOD and offers a practical path toward fine-grained grounding-based open-vocabulary detectors.
\end{abstract}

\begin{IEEEkeywords}
Open-vocabulary object detection, fine-grained recognition, vision-language models, sentence-level grounding.
\end{IEEEkeywords}
\section{Introduction}
\label{sec:intro}

Open-vocabulary object detection (OVOD) enables detectors to recognize categories specified by natural-language text rather than by a fixed classifier. This flexibility is important in practical domains such as vehicle perception and product recognition, where new models, variants, or fine-grained categories may appear after deployment. Among OVOD models, grounding-based detectors such as GLIP and Grounding DINO~\cite{li2022grounded,liu2024grounding} are particularly attractive because they provide a widely adopted text-conditioned detection pipeline with deep cross-modal fusion and strong localization capability. 
However, these detectors were originally designed for phrase grounding and referring expressions, where token-level alignment is natural because individual words may correspond to objects, attributes, or relations. For fine-grained category prediction, this interface becomes indirect because the target is not a phrase span to be localized but a category identity defined by a complete description.
Our goal is therefore to preserve the deep-fusion grounding detector while reformulating its text-side interface for fine-grained recognition.


\begin{figure}[t]
  \centering
  \includegraphics[width=\linewidth]{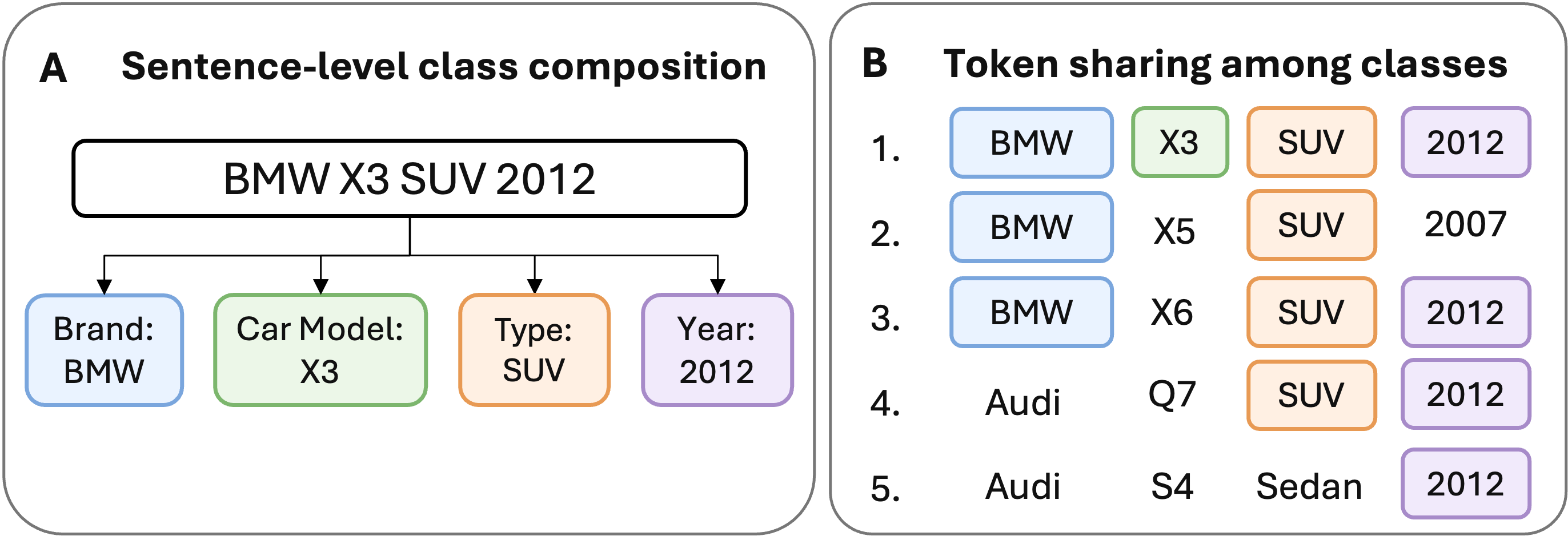}
 \caption{Example of compositional fine-grained labels and token sharing in vehicle categories. 
 }
  \label{fig:class_composition}

  \vspace{-5mm}
\end{figure}

As illustrated in Fig.~\ref{fig:class_composition}, fine-grained vehicle categories provide a concrete object-level detection example that exposes this mismatch. A Stanford Cars category name~\cite{krause2013object}, such as \textit{BMW X3 SUV 2012}, contains multiple textual components, including brand, model name, vehicle type, and model year. These components play different roles in category semantics: the vehicle type is visually descriptive and relatively coarse, whereas model name fragments such as \textit{X3} or \textit{S4} are visually under-specified when treated in isolation. This motivates the use of complete category descriptions rather than isolated model name tokens. Moreover, different candidate categories may share the same brand, type, or other tokens, as in \textit{BMW X3 SUV} and \textit{BMW X5 SUV}. This structure allows us to examine fine-grained OVOD at multiple label granularities. We evaluate type, brand, and car model recognition, where the car model level is represented by the complete composition of brand, model name, and vehicle type. 
These controlled granularities let us test whether the token-level interface degrades more severely as category identity becomes finer and more compositional.

To address this issue, we propose SL-Ground, a sentence-level grounding formulation for Grounding DINO-style detectors. Instead of serializing all candidate categories into a token sequence, SL-Ground encodes each candidate category prompt independently with a CLIP text encoder~\cite{radford2021learning}. The resulting sentence-level embedding serves as the text representation of each category, and object queries are scored directly against these sentence-level features to produce query-to-class logits. We instantiate SL-Ground in a Grounding DINO-based query detector~\cite{liu2024grounding}, keeping the standard detection pipeline while replacing the text-side scoring interface. To isolate the effect of sentence-level grounding from the text encoder choice, we compare SL-Ground with a CLIP-based token-level variant under the same detector pipeline. Beyond accuracy, this formulation also improves inference scalability: token-level grounding depends on the total token length of the serialized vocabulary and may require repeated prompt chunking as the category set grows, whereas SL-Ground represents each category as an independent class feature and avoids repeated token-level prompt chunking.

Controlled experiments on Stanford Cars show that under the same training candidate-sampling policy as the token-level variant, SL-Ground raises car model AP from 14.5 to 70.6 under the same detector pipeline, and full-vocabulary training further reaches 86.0. Further evaluations show better preserved open-vocabulary performance after fine-grained adaptation, competitive COCO and miniLVIS results, improved inference scalability, and the best average AP on the FG-OVD benchmark, reaching 61.6. 
Our contributions are summarized as follows:

\begin{itemize}

\item We introduce a multi-granularity fine-grained OVOD protocol, covering vehicle type, brand, and car model labels, and use it to identify token-level grounding as a key bottleneck for query-based grounding detectors.

\item We propose SL-Ground, a sentence-level grounding formulation that replaces serialized token-level category representations with one sentence-level embedding per category for direct query-to-class prediction.

\item We conduct extensive experiments showing that SL-Ground significantly improves car model detection and also generalizes to broader fine-grained recognition, better preserves open-vocabulary performance after adaptation, and scales more efficiently by avoiding repeated token-level prompt chunking.

\end{itemize}

\vspace{-3mm}
\section{Related Work}
\label{sec:related_works}

\subsection{Open-Vocabulary Object Detection}

Open-vocabulary object detection (OVOD) addresses the fixed-category limitation of conventional detectors by representing categories with natural-language descriptions or text embeddings. Early work, such as Zareian et al.~\cite{zareian2021open}, established the base-to-novel detection setting, while CLIP~\cite{radford2021learning} provided transferable image-text representations for later OVOD methods. ViLD~\cite{gu2022open}, F-VLM~\cite{kuo2023open}, RegionCLIP~\cite{zhong2022regionclip}, and Detic~\cite{zhou2022detecting} transfer vision-language knowledge through distillation, frozen vision-language models, region-level pretraining, or image-level supervision. OWL-ViT~\cite{minderer2022simple} and OWLv2~\cite{minderer2023scaling} match independently encoded category prompts to regions. DetCLIP~\cite{yao2022detclip} encodes concepts independently through a paralleled formulation rather than concatenating category names into a single prompt. DetCLIPv2~\cite{yao2023detclipv2} extends this approach to large-scale pretraining through word-region alignment.

Grounding-based detectors integrate language into the detection pipeline through cross-modal fusion. GLIP~\cite{li2022grounded} and Grounding DINO~\cite{liu2024grounding} align regions with category names or referring expressions and fuse visual and textual features before final prediction. Grounding DINO further employs language-guided query selection and a cross-modal decoder. Independently encoded category representations can also participate in vision-language fusion. YOLO-World~\cite{cheng2024yoloworld} combines CLIP text embeddings with visual features through RepVL-PAN. OV-DINO~\cite{wang2024ovdino} uses mean-pooled BERT representations for category prompts and language-aware query selection, followed by gated cross-attention between decoder queries and the selected visual features. Recent methods such as DINO-X~\cite{ren2024dinox} and LLMDet~\cite{fu2025llmdet} further develop open-vocabulary detection through large-scale object-centric training or LLM-generated captions.

Our work examines how the grounding unit affects discrimination among compositional fine-grained categories. The original Grounding DINO formulation matches object queries to individual text tokens and recovers class scores through token-to-class aggregation. SL-Ground replaces these token representations with independently encoded CLIP sentence features. Unlike OV-DINO's selective fusion of visual features, SL-Ground retains Grounding DINO's bidirectional image-text fusion and direct query-to-text attention, allowing complete category descriptions to participate directly in feature fusion and query refinement. We study this formulation using a CLIP-based token-level controlled variant and matched training candidate sampling to assess whether the improvement in fine-grained discrimination extends beyond text encoder choice and the number of training candidates.

\subsection{Fine-Grained Visual Recognition}

Fine-grained visual recognition distinguishes subordinate categories
within a shared basic-level concept, such as bird
species~\cite{wah2011caltech}, car
models~\cite{krause2013object,tafazzoli2017large}, and dog
breeds~\cite{khosla2011novel}. Because these categories often differ
only in subtle local details, prior methods exploit localized
discriminative evidence through part-aware
CNNs~\cite{lin2015bilinear,fu2017look} or transformer-based token
selection~\cite{he2022transfg}. Vehicle recognition is particularly
relevant to our setting because it spans multiple label granularities,
from vehicle types and brands to specific car models, while requiring
discrimination among visually similar categories. Supervised vehicle
recognition methods further improve performance with global context,
local evidence, or attention to diagnostic
regions~\cite{amirkhani2023deepcar,tan2023coarse,hassan2021empirical}.
However, most fine-grained recognition methods assume a closed
vocabulary and require additional annotation or retraining for new
categories. Fine-grained distinctions also remain challenging under
open-set recognition~\cite{lang2024coarse}, motivating open-vocabulary
fine-grained detection beyond closed-set recognition.

\subsection{Fine-Grained Vision-Language Semantics}

Recent work has examined fine-grained semantics in vision-language
models from evaluation, prompt design, and representation
perspectives. FG-OVD~\cite{bianchi2024devil} introduces an
attribute-level hard-negative benchmark and shows that open-vocabulary
detectors often struggle with semantically similar prompts.
HA-FGOVD~\cite{ma2024hafgovd}, GUIDED~\cite{li2026guided}, and
DSAA~\cite{jiang2026dsaa} illustrate this direction by adding
attribute-oriented components on top of Grounding DINO-style
token-level detection pipelines. These works primarily study
distinctions expressed through visual attributes such as color,
material, texture, pattern, and transparency. Other studies improve or
evaluate fine-grained semantics from complementary angles.
BIND~\cite{zhang2024exploring} studies region-word alignment before
detection training, FG-CLIP~\cite{xie2025fgclip} improves CLIP
fine-grained alignment with long captions, region-level annotations,
and hard negatives, and SHiNe~\cite{liu2024shine} uses hierarchy-aware
sentence embeddings for open-vocabulary classifiers.
3F-OVD~\cite{liu2025finegrained} introduces fine-grained prompts and a
large-scale vehicle and retail-product benchmark.
Finedefics~\cite{he2025analyzing} studies subordinate recognition in
multimodal large language models.

SL-Ground represents each category as a complete compositional
description rather than an attribute-rich prompt over a shared object
category. Rather than highlighting individual attribute tokens or
adding an attribute-modeling module, we make a minimal change to the
grounding interface, representing each fine-grained category as a
single sentence-level embedding so that the category is scored
directly on the class axis rather than recovered from token-level
evidence.
\section{Proposed Methods}
\label{sec:methods}
This section presents SL-Ground as a modification to the language-side grounding interface rather than a new detection pipeline. The fine-grained open-vocabulary detection setting and the compositional structure of class descriptions are first formalized to establish the notation shared by the token-level formulation and our method (Sec.~\ref{sub:problem_formulation}). Token-level encoding is then revisited as the reference formulation (Sec.~\ref{sub:preliminary}), providing a direct contrast to the proposed sentence-level grounding architecture (Sec.~\ref{sub:sentence_level_grounding}). Finally, Sec.~\ref{sub:training_objective} describes the training objective.

\subsection{Problem Formulation}
\label{sub:problem_formulation}

Given an input image \(I\) and a user-provided open-vocabulary class set
\begin{equation}
  \mathcal{C} = \{c_1,\ldots,c_K\},
\end{equation}
an open-vocabulary object detector predicts a set of localized object instances
\begin{equation}
  \mathcal{Y}(I,\mathcal{C}) = \{(b_i, y_i, s_i)\}_{i=1}^{N},
\end{equation}
where \(b_i\) denotes a predicted bounding box, \(y_i \in \mathcal{C}\) is the predicted class label, and \(s_i\) is the corresponding confidence score. Unlike closed-set detection, the vocabulary \(\mathcal{C}\) is specified by arbitrary text descriptions and can be extended to unseen categories at inference time without retraining or replacing the classifier. 

Each fine-grained class \(c_k\) can be represented by a compositional description
\begin{equation}
  c_k =
  \left(
  a_k^{(1)}, a_k^{(2)}, \ldots, a_k^{(M_k)}
  \right),
\end{equation}
where each \(a_k^{(m)}\) denotes a semantic component of the target class. In practice, the text encoder input is formulated as
\begin{equation}
  q_k = p(c_k),
  \label{eq:category_prompt}
\end{equation}
where \(p(\cdot)\) either keeps the original class phrase or places it into a prompt template, e.g., \textit{``BMW X3 SUV''} or \textit{``a photo of a BMW X3 SUV''}. We use a fixed prompt template across all variants to isolate the effect of the grounding formulation. 
\subsection{Preliminary: Token-Level Encoding}
\label{sub:preliminary}

Grounding-based OVOD detectors commonly formulate text-conditioned detection as token-level region-text alignment. Given a category vocabulary \(\mathcal{C}\), the category names or phrases are serialized into a joint text prompt and encoded into token-level text embeddings
\begin{equation}
  \mathbf{T} \in \mathbb{R}^{B \times L \times D},
\end{equation}
where \(B\) is the batch size, \(L\) is the maximum text sequence length of the text encoder, and \(D\) is the hidden dimension. 
The similarity logits between individual queries and tokens could be represented as
\begin{equation}
  \mathbf{S}^{\mathrm{tok}} \in \mathbb{R}^{B \times Q \times L},
\end{equation}
where \(Q\) is the number of object queries.

To identify objects described by a sequence of tokens, the final similarity scores must be recovered from token-level logits. Let \(\mathcal{A}_k \subseteq \{1,\ldots,L\}\) denote the valid token indices associated with the phrase of class \(c_k\). The score of class \(c_k\) for the \(j\)-th query is obtained by
\begin{equation}
  s_{j,k} =
  \mathrm{Agg}_{\ell \in \mathcal{A}_k}
  \mathbf{S}^{\mathrm{tok}}_{j,\ell},
  \label{eq:token_to_class_aggregation}
\end{equation}
where \(\mathrm{Agg}(\cdot)\) is a token-to-class aggregation operation, such as averaging or taking the maximum over the associated tokens.
This formulation suffers from two major drawbacks. First, the final score is derived from the aggregation operation, which may fail to preserve the complete semantic information contained in the original description. Second, the computational cost is highly correlated with the total token length processed in the feed-forward pass.

\begin{figure*}[t]
  \centering
  \includegraphics[
    width=0.98\textwidth,
    keepaspectratio
  ]{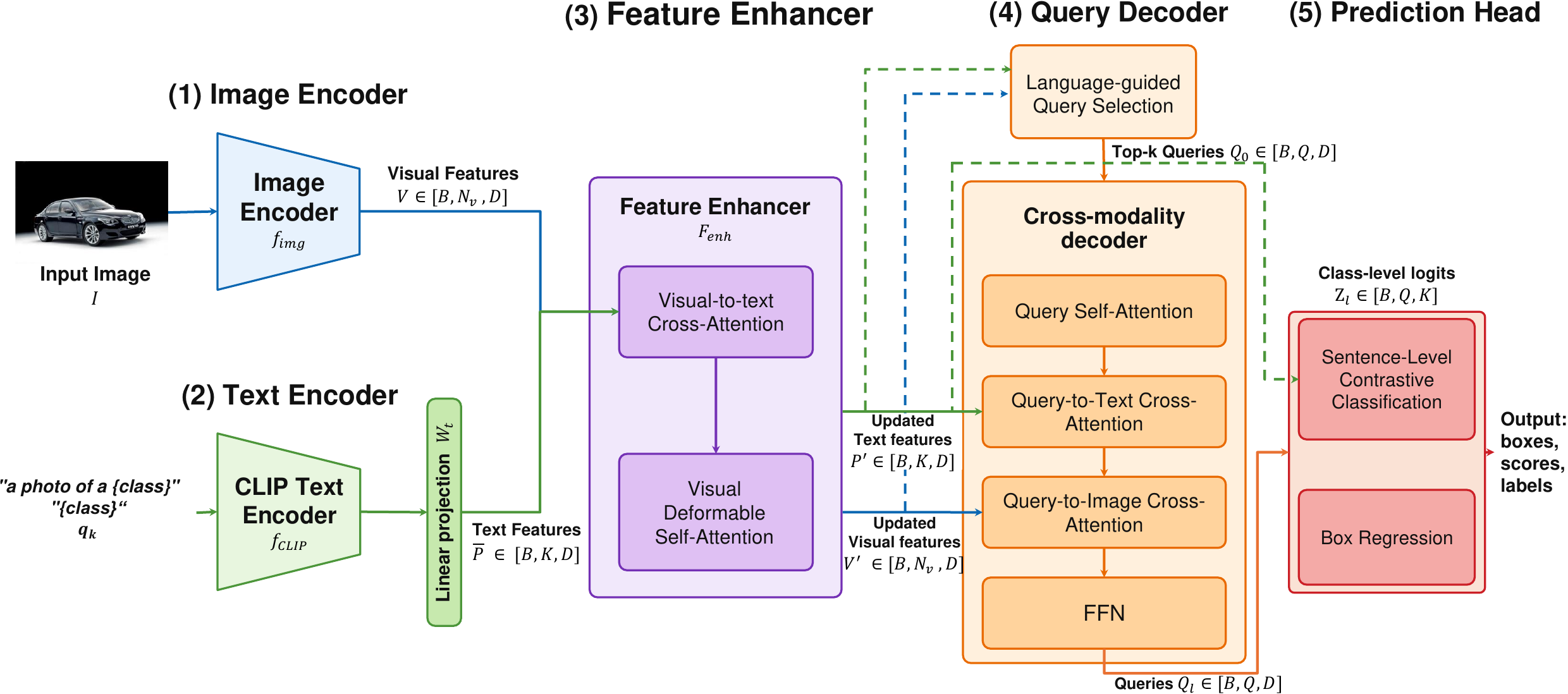}
\caption{Overview of SL-Ground. Each class description is independently encoded into one CLIP sentence feature. The feature enhancer fuses visual and class features, class-level proposal scores initialize the decoder, and the prediction outputs boxes and query-to-class logits.}
  \label{fig:groundingclip_architecture}
\end{figure*}

\subsection{Sentence-Level Grounding}
\label{sub:sentence_level_grounding}


SL-Ground builds on the core architecture of Grounding DINO~\cite{liu2024grounding}, but changes the text representation used for grounding. Instead of serializing category descriptions into token sequences, SL-Ground encodes each candidate category as a sentence-level class feature and carries this class axis through cross-modal fusion, proposal selection, decoder attention, and final query-to-class prediction. By avoiding the token-to-class aggregation in \eqref{eq:token_to_class_aggregation}, SL-Ground differs from a text-encoder replacement alone, which would still recover fine-grained class scores from token-level responses.


The proposed architecture is illustrated in Fig.~\ref{fig:groundingclip_architecture}, and its components are presented in detail in the following.
\subsubsection{Image encoder}
Given an input image \(I\), the image encoder extracts hierarchical visual features, applies a feature neck
and projection layers, and converts the resulting multi-scale feature maps into a sequence of visual tokens:
\begin{equation}
  \mathbf{V}
  =
  f_{\mathrm{img}}(I)
  \in
  \mathbb{R}^{B \times N_v \times D},
\end{equation}
Here, \(f_{\mathrm{img}}\) denotes the image encoder, \(N_v\) is the total number of visual tokens, and \(\mathbf{V}\) is the resulting visual token sequence.

\subsubsection{Sentence-level text encoder}
For each candidate category \(c_k \in \mathcal{C}\), we construct a textual description \(q_k\) as defined in \eqref{eq:category_prompt}. Instead of serializing all class descriptions into one token sequence, each description is processed as a separate CLIP text input. We use the pooled feature at the end-of-text (EOT) position as the sentence-level class representation and project it into the detector hidden dimension:
\begin{equation}
  \bar{\mathbf{p}}_k
  =
  W_t f_{\mathrm{CLIP}}(q_k)_{\mathrm{EOT}}
  \in
  \mathbb{R}^{D}.
\end{equation}
Here, \(W_t\) denotes a text projection layer that maps the CLIP sentence feature into the detector hidden dimension. The \(K\) candidate categories are then represented by the class feature matrix
\begin{equation}
  \bar{\mathbf{P}} =
  [\bar{\mathbf{p}}_1,\ldots,\bar{\mathbf{p}}_K]^\top
  \in \mathbb{R}^{K \times D}.
\end{equation}
Here, \(\bar{\mathbf{P}}\) stacks the projected sentence-level class features as rows and serves as the text-side class representation for the detector. This changes the text-side axis from \(L\) token embeddings to \(K\) class embeddings, so query-to-class scoring is computed over \([B,Q,K]\) rather than recovered from token-level logits by aggregation.

\subsubsection{Feature enhancer}
The class feature matrix \(\bar{\mathbf{P}}\) is broadcast across the batch and
fused with the visual tokens by the feature enhancer:
\begin{equation}
  (\mathbf{V}',\mathbf{P}')
  =
  F_{\mathrm{enh}}(\mathbf{V},\bar{\mathbf{P}}).
\end{equation}
Here, \(F_{\mathrm{enh}}\) denotes the feature enhancer, which performs cross-modality feature fusion between the visual tokens and sentence-level class features. \(\mathbf{V}' \in \mathbb{R}^{B \times N_v \times D}\) and \(\mathbf{P}' \in \mathbb{R}^{B \times K \times D}\) denote the updated visual and class features, respectively.

Since the CLIP text encoder produces a sentence-level representation for each input description, the original design that uses self-attention and masking to extract sub-sentence-level text features is no longer necessary. We therefore remove this sublayer while retaining the bidirectional image-text fusion layers. This preserves cross-modal interactions between sentence-level textual semantics and visual object features, while keeping the detector-side text axis tied to \(K\) classes rather than token length \(L\).

\subsubsection{Class-level query selection}
The updated visual tokens are compared directly with the updated class features
to produce encoder proposal logits:
\begin{equation}
  \mathbf{S}_{\mathrm{enc}}^{\mathrm{cls}}
  =
  \frac{\mathbf{V}'{\mathbf{P}'}^\top}{\sqrt{D}},
  \quad
  \mathbf{S}_{\mathrm{enc}}^{\mathrm{cls}}
  \in \mathbb{R}^{B \times N_v \times K}.
\end{equation}
For each visual position, its proposal score is obtained from the highest
candidate-class response:
\begin{equation}
  r_{b,n}
  =
  \max_{1 \leq k \leq K}
  \left(
    \mathbf{S}_{\mathrm{enc}}^{\mathrm{cls}}
  \right)_{b,n,k}.
\end{equation}
The \(Q\) visual positions with the highest proposal scores are selected to
initialize the object queries for the decoder.

\subsubsection{Query decoder and prediction head}
The selected proposals form the initial object queries
\begin{equation}
  \mathbf{Q}_0
  \in
  \mathbb{R}^{B \times Q \times D}.
\end{equation}
At each decoder layer, the queries are refined through query self-attention,
query-to-text cross-attention over \(\mathbf{P}'\), query-to-image deformable
attention over \(\mathbf{V}'\), and feed-forward layers. Let
\(\mathbf{Q}_{\ell}\) denote the resulting query features at decoder layer
\(\ell\).

The prediction head computes a score for each query and candidate class:

\begin{equation}
  z^{\ell}_{b,q,k}
  =
  \frac{
    \left\langle
      \mathbf{q}^{\ell}_{b,q},
      \mathbf{p}'_{b,k}
    \right\rangle
  }{\sqrt{D}}
  +
  \gamma_{\ell},
\end{equation}
where \(\gamma_{\ell}\) is a learnable scalar bias added to the query-to-class logits. The resulting
class-level logit tensor is
\begin{equation}
  \mathbf{Z}_{\ell}
  \in
  \mathbb{R}^{B \times Q \times K}.
\end{equation}
The box regression branch remains unchanged and predicts one bounding box from
each decoder query.


With the proposed method, sentence-level semantic representations are used throughout the detection pipeline, including cross-modal fusion, proposal selection, decoder attention, and final prediction. By avoiding the aggregation process required in token-level grounding, the scoring unit remains aligned with the complete category description, which is especially important when fine-grained classes share individual tokens.
\vspace{-2mm}

\subsection{Training Objective}
\label{sub:training_objective}

We train the detector with the standard set prediction objective, including classification, box regression, and generalized IoU losses:
\[
\mathcal{L}
=
\lambda_{\mathrm{cls}}\mathcal{L}_{\mathrm{cls}}
+
\lambda_{1}\mathcal{L}_{1}
+
\lambda_{\mathrm{giou}}\mathcal{L}_{\mathrm{giou}}.
\]
The classification loss is applied directly to query-to-class logits computed from sentence-level class features. This differs from token-level grounding, where classification logits are defined over text tokens and class scores are recovered through token-to-class aggregation.
\vspace{-3mm}

\section{Experimental Results}
\label{sec:exp}


We first evaluate the controlled Grounding DINO-based variants on Stanford Cars (Sec.~\ref{sub:exp_stanfordcar}), examine cross-granularity transfer and OVOD preservation after adaptation (Sec.~\ref{sub:transfer_retention}), and report standard OVOD results (Sec.~\ref{sub:ovod_sota}). We then examine performance beyond the Stanford Cars detection setting on FG-OVD (Sec.~\ref{sub:fgovd}) and image-level fine-grained recognition (Sec.~\ref{sub:exp_finegrained}). These benchmark results are followed by a control for training candidate set size on Stanford Cars (Sec.~\ref{sub:training_candidate_size}), and by analyses of class separation in the projected embedding space (Sec.~\ref{sub:scoring_space}) and inference efficiency (Sec.~\ref{sub:FPS}). Finally, qualitative comparisons illustrate class confusion and prompt-chunking behavior (Sec.~\ref{sub:qualitative}).

\subsection{Experimental Setup}
\label{sub:experimental_setup}

\subsubsection{Datasets}
We evaluate standard open-vocabulary detection on COCO~\cite{lin2014microsoft} and miniLVIS~\cite{gupta2019lvis}, fine-grained open-vocabulary detection on FG-OVD~\cite{bianchi2024devil}, and Stanford Cars~\cite{krause2013object}. For Stanford Cars, we remove the model year from each of the 196 original class names and merge classes with identical remaining names, resulting in 189 car-model categories. We evaluate three label granularities: type, brand, and car model. The car model label combines brand, model name, and vehicle type. For example, ``Suzuki Aerio Sedan 2007'' becomes the car model label ``Suzuki Aerio Sedan'', with ``Suzuki'' and ``Sedan'' used as its brand and type labels.


\subsubsection{Compared Methods}
We compare SL-Ground with controlled Grounding DINO-based variants and published OVOD baselines. Under the same Swin-T~\cite{liu2021swin} detection pipeline, we evaluate \textit{GD-BERT}, the original Grounding DINO formulation with BERT token-level grounding; \textit{GD-CLIP}, a CLIP text-encoder variant that preserves token-level grounding; and SL-Ground, a CLIP sentence-level grounding. This setup isolates the effects of the text encoder and grounding granularity. For published baselines, we report AP from the original papers or official result tables when available.

\subsubsection{Implementation Details}
All controlled variants are initialized from Objects365v1 (O365v1)~\cite{shao2019objects365} pretraining and use the same prompt template and detection pipeline. O365v1 pretraining uses AdamW for 30 epochs with learning rate \(3\times10^{-4}\), weight decay \(1\times10^{-4}\), and batch size 16 per GPU. Stanford Cars fine-tuning uses AdamW for 20 epochs with learning rate \(1\times10^{-4}\), weight decay \(1\times10^{-4}\), and batch size 8 per GPU. Token-level variants sample candidate classes per image within the text encoder length limit, whereas SL-Ground trains with the full class vocabulary by default (Sec.~\ref{sub:training_candidate_size}). For FG-OVD, SL-Ground is fine-tuned for 2000 iterations following GUIDED~\cite{li2026guided}. Efficiency is measured on a single RTX 4090 with text encoding included and no text-feature caching; token-level baselines re-encode each prompt chunk when chunking is required.

\subsection{Fine-Grained Detection on Stanford Cars}
\label{sub:exp_stanfordcar}

To examine the effects of the text encoder and grounding interface, Table~\ref{tab:stanfordcar_main} evaluates Stanford Cars detection at three label granularities using the same core detection pipeline, optimization settings, and evaluation protocol. Replacing BERT with CLIP examines the role of image-text contrastive pretraining in vehicle label representations, while the comparison between CLIP token-level grounding and query-to-class scoring examines whether fine-grained detection is further limited by the token-based grounding formulation.

\begin{table}[t]
  \centering
\caption{Stanford Cars detection results across label granularities.}  
\label{tab:stanfordcar_main}
  \resizebox{\columnwidth}{!}{%
  \begin{tabular}{lccccc}
    \toprule
    \multicolumn{3}{c}{Detector setting} & 
    \multicolumn{3}{c}{Label granularity (AP $\uparrow$)} \\
    \cmidrule(lr){1-3} \cmidrule(lr){4-6}
    Model & \makecell{Text\\encoder} & \makecell{Grounding\\unit} 
    & Type & Brand & Car model \\
    \midrule
    GD-BERT & BERT & Token & 70.3 & 40.1 & 10.3 \\
    GD-CLIP & CLIP & Token & 80.5 & 73.3 & 14.5 \\
    \midrule
    SL-Ground  & CLIP & Sentence 
    & \textbf{88.0} & \textbf{92.8} & \textbf{86.0} \\
    \bottomrule
  \end{tabular}
  }
  \par\smallskip
 {\footnotesize\raggedright
Each result column uses the corresponding label granularity for both fine-tuning and evaluation. Token-level variants sample training candidates when the class vocabulary exceeds their prompt length limit, whereas SL-Ground uses the full class vocabulary (Sec.~\ref{sub:training_candidate_size}).\par}
\end{table}

We compare the text encoder and grounding formulation for fine-grained open-vocabulary detection. Replacing BERT with CLIP while retaining token-level grounding (GD-BERT $\rightarrow$ GD-CLIP) improves type AP from 70.3 to 80.5 and brand AP from 40.1 to 73.3, suggesting that image-text pretraining provides text features that are more compatible with visual vehicle semantics. At the car model granularity, however, the same change yields only a modest gain from 10.3 to 14.5 AP. With the CLIP encoder fixed, adopting sentence-level grounding throughout fusion, query selection, and prediction (GD-CLIP $\rightarrow$ SL-Ground) increases car model AP from 14.5 to 86.0. This comparison supports the benefit of the sentence-level grounding formulation beyond the text encoder choice; it does not isolate the contribution of each detector-side change. The two variants also differ in training candidate exposure, since GD-CLIP is trained on classes sampled under its 75-token prompt limit, whereas SL-Ground is trained with all 189 car model classes. Under the same candidate-sampling policy, SL-Ground still reaches 70.6 AP (Sec.~\ref{sub:training_candidate_size}).

\begin{figure}[t]
  \centering
  \subfloat[Brand]{\includegraphics[width=0.495\linewidth]{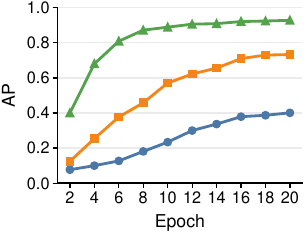}}%
  \subfloat[Car model]{\includegraphics[width=0.495\linewidth]{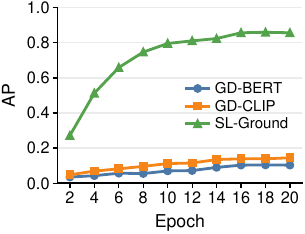}}
 \caption{Epoch-wise Stanford Cars AP over 20 fine-tuning epochs. SL-Ground stays above the token-level variants throughout, with the largest gap at car model granularity.}
  \label{fig:stanfordcar_epoch_map}

\end{figure}

Fig.~\ref{fig:stanfordcar_epoch_map} shows the AP training history in the granularities of  brand and car model. While CLIP token-level grounding improves brand-level learning, SL-Ground consistently converges to higher AP. At the car model level, both token-level variants remain low throughout fine-tuning, whereas SL-Ground separates early and reaches substantially higher AP. This suggests that this performance gap
 is not simply a difference in convergence rates, but persists under the evaluated training schedule.

\subsection{Effects of Fine-Grained Adaptation}
\label{sub:transfer_retention}

\subsubsection{Cross-Granularity Transfer}
\label{sub:cross-granularity}

The Stanford Cars labels expose three label granularities: type, brand, and car model. Same-granularity evaluation measures the target task, while off-diagonal entries test whether a model fine-tuned for one granularity transfers to another. The columns labeled Type, Brand, and Car model in Table~\ref{tab:combined_granularity_generalization} report this transfer matrix.

The diagonal entries show that SL-Ground is most effective when the fine-tuning and evaluation label granularities match, reaching 88.0 AP for type, 92.8 AP for brand, and 86.0 AP for car model. The off-diagonal entries, however, indicate that cross-granularity transfer is limited. Models trained on type labels achieve high type AP, but remain near zero on brand and car model evaluation, showing that coarse body-type supervision does not provide enough information to recover identity-level labels. Training on brand labels provides only partial transfer to car model recognition, with the best off-diagonal car model AP reaching 13.7.

The car model training rows further show that off-diagonal transfer should not be interpreted as fine-grained understanding. GD-BERT and GD-CLIP obtain nontrivial type or brand AP after car model fine-tuning, but their same-granularity car model AP remains low at 10.3 and 14.5. In contrast, SL-Ground strongly optimizes the intended car model label space, reaching 86.0 AP, but does not automatically decompose this representation into accurate type or brand predictions. These results suggest that the three granularities behave as related but distinct label spaces rather than a simple hierarchy, supporting the need to preserve complete class-level descriptions for fine-grained detection.

\begin{table}[t]
  \centering
  \caption{Cross-granularity transfer and OVOD knowledge preservation after Stanford Cars fine-tuning. Rows indicate the fine-tuning target, and COCO/miniLVIS AP measure retained open-vocabulary performance.}
  \label{tab:combined_granularity_generalization}
  \resizebox{\columnwidth}{!}{%
  \begin{tabular}{llccccc}
    \toprule
    \multicolumn{2}{c}{Fine-tuning setting} &
    \multicolumn{3}{c}{Stanford Cars AP $\uparrow$} &
    \multicolumn{2}{c}{OVOD AP $\uparrow$} \\
    \cmidrule(lr){1-2} \cmidrule(lr){3-5} \cmidrule(lr){6-7}
    \makecell{Fine-tune\\target} & Model
    & Type & Brand & \makecell{Car\\model}
    & COCO & miniLVIS \\
    \midrule

    \multirow{3}{*}{Zero-shot}
      & GD-BERT   & 9.6 & 1.0 & 0.3 & 48.4 & 15.78 \\
      & GD-CLIP   & 1.6 & 0.6 & 0.1 & 45.6 & \textbf{26.37} \\
      & SL-Ground & 1.4 & 0.6 & 0.9 & \textbf{48.8} & 25.27 \\
    \midrule

    \multirow{3}{*}{Type}
      & GD-BERT   & 70.3 & 3.0 & 2.6 & 15.8 & 6.01 \\
      & GD-CLIP   & 80.5 & 3.6 & 5.1 & 21.2 & 11.36 \\
      & SL-Ground & \textbf{88.0} & 0.6 & 4.2 & \textbf{24.9} & \textbf{12.70} \\
    \midrule

    \multirow{3}{*}{Brand}
      & GD-BERT   & 10.0 & 40.1 & 7.4 & 19.4 & 5.14 \\
      & GD-CLIP   & 9.1 & 73.3 & 7.9 & 15.5 & 7.18 \\
      & SL-Ground & 1.2 & \textbf{92.8} & 13.7 & \textbf{30.2} & \textbf{12.76} \\
    \midrule

    \multirow{3}{*}{\makecell{Car\\model}}
      & GD-BERT   & 31.5 & 6.8 & 10.3 & 16.1 & 4.70 \\
      & GD-CLIP   & 28.6 & 17.2 & 14.5 & 9.6 & 7.27 \\
      & SL-Ground & 5.8 & 6.9 & \textbf{86.0} & \textbf{30.5} & \textbf{12.62} \\
    \bottomrule
  \end{tabular}%
  }

  \vspace{-5mm}
\end{table}

\subsubsection{OVOD Knowledge Preservation}
\label{sub:forgetting}
Beyond target-dataset AP, we evaluate whether each fine-tuned checkpoint retains general open-vocabulary detection performance on COCO and miniLVIS, as reported in the last two columns of Table~\ref{tab:combined_granularity_generalization}. Because the text encoder is frozen during fine-tuning, any drop in preserved OVOD performance reflects adaptation of the detector-side parameters rather than text-encoder drift.

The token-level variants preserve substantially less OVOD knowledge, especially after car model fine-tuning. GD-BERT and GD-CLIP retain only 16.1 and 9.6 COCO AP, and 4.70 and 7.27 miniLVIS AP, respectively. At the same time, their target car model AP remains low at 10.3 and 14.5, indicating that fine-grained adaptation neither solves the intended car model task nor preserves broad open-vocabulary performance well.

In contrast, SL-Ground achieves 86.0 car model AP while retaining 30.5 COCO AP and 12.62 miniLVIS AP under the same fine-tuning setting. Although fine-tuning still degrades OVOD knowledge preservation compared with the zero-shot checkpoint, SL-Ground consistently retains higher COCO and miniLVIS AP than the token-level variants. This suggests that representing each category with an independently encoded sentence-level CLIP embedding provides a more stable open-vocabulary anchor during fine-grained adaptation.

\begin{table}[t]
  \centering
  \caption{COCO open-vocabulary detection results.}
  \label{tab:sota_coco}
  \resizebox{\columnwidth}{!}{%
  \begin{tabular}{llcc}
    \toprule
    Method & Pretrain & Data (M) & AP $\uparrow$ \\
    \midrule
    GLIP~\cite{li2022grounded} & O+G+C4\textsuperscript{\dag} & 5.43 & 46.3 \\
    Grounding DINO-T(3)~\cite{liu2024grounding} & O+G+C4\textsuperscript{\dag} & 5.43 & 48.4 \\
    YOLO-World-L~\cite{cheng2024yoloworld} & O+G+C\textsuperscript{\dag} & 1.63 & 45.1 \\
    Dynamic-DINO~\cite{lu2025dynamicdino} & O+G+V\textsuperscript{\dag} & 1.56 & 46.2 \\
    \midrule
    GLIP-T(C)~\cite{li2022grounded} & O+G\textsuperscript{\dag} & 1.43 & 46.7 \\
    Grounding DINO-T(2)~\cite{liu2024grounding} & O+G\textsuperscript{\dag} & 1.43 & 48.1 \\
    YOLO-World-S~\cite{cheng2024yoloworld} & O+G\textsuperscript{\dag} & 1.43 & 37.6 \\
    YOLO-World-M~\cite{cheng2024yoloworld} & O+G\textsuperscript{\dag} & 1.43 & 42.8 \\
    OmDet-Turbo-T~\cite{zhao2024omdetturbo} & O+G\textsuperscript{\dag} & 1.43 & 42.5 \\
    YOLOE-v8~\cite{wang2025yoloe} & O+G\textsuperscript{\dag} & 1.43 & 42.0 \\
    YOLOE-v11~\cite{wang2025yoloe} & O+G\textsuperscript{\dag} & 1.43 & 41.7 \\
    OWL-ViT~\cite{minderer2022simple} & O+VG\textsuperscript{\dag} & 0.77 & 42.2 \\
    \midrule
    GLIP-T(A)~\cite{li2022grounded} & O\textsuperscript{*} & 0.66 & 42.9 \\
    GLIP-T(B)~\cite{li2022grounded} & O\textsuperscript{*} & 0.66 & 44.9 \\
    Grounding DINO-T(1)~\cite{liu2024grounding} & O\textsuperscript{*} & 0.66 & 46.7 \\
    OV-DINO~\cite{wang2024ovdino} & O\textsuperscript{*} & 0.60 & \textbf{49.5} \\
    Grounding DINO-1.5Edge~\cite{ren2024groundingdino15} & G20M\textsuperscript{\ddag} & 20 & 45.0 \\
    \midrule
    SL-Ground & O\textsuperscript{*} & 0.66 & 48.8 \\
    \bottomrule
  \end{tabular}%
  }
  \par\smallskip
  {\footnotesize\raggedright
  \textsuperscript{*}O denotes Objects365. 
  \textsuperscript{\dag}Additional sources include G=GoldG, C4=Cap4M, C=CC3M, V=V3Det, and VG=Visual Genome. 
  \textsuperscript{\ddag}G20M denotes Grounding-20M. Data sizes follow the source papers.\par}
  \vspace{-3mm}
\end{table}

\begin{table}[t]
  \centering
  \caption{miniLVIS open-vocabulary detection results.}
  \label{tab:sota_minilvis}
  \resizebox{\columnwidth}{!}{%
  \begin{tabular}{lcccc}
    \toprule
    Method & AP $\uparrow$ & AP$_r$ $\uparrow$ & AP$_c$ $\uparrow$ & AP$_f$ $\uparrow$ \\
    \midrule
    GLIP-T(A)~\cite{li2022grounded} & 18.5 & 14.2 & 13.9 & 23.4 \\
    GLIP-T(B)~\cite{li2022grounded} & 17.8 & 13.5 & 12.8 & 22.2 \\
    YOLO-World-L~\cite{cheng2024yoloworld} & 23.5 & 16.2 & 21.1 & 27.0 \\
    Grounding DINO-T~\cite{liu2024grounding} & 16.0 & 5.0 & 10.0 & 23.0 \\
    OV-DINO~\cite{wang2024ovdino} & 24.4 & 15.5 & 20.2 & \textbf{29.7} \\
    \midrule
    SL-Ground & \textbf{25.27} & \textbf{18.52} & \textbf{21.84} & 29.52 \\
    \bottomrule
  \end{tabular}%
  }
  \par\smallskip
  {\footnotesize\raggedright
  All methods are pretrained on O365 only (O\textsuperscript{*} in Table~\ref{tab:sota_coco}). AP$_r$, AP$_c$, and AP$_f$ denote rare, common, and frequent category AP.\par}
  \vspace{-5mm}
\end{table}
\subsection{Standard OVOD Benchmarks}
\label{sub:ovod_sota}

We evaluate SL-Ground on standard COCO and miniLVIS benchmarks to verify that sentence-level grounding remains competitive beyond fine-grained category grounding. Since SL-Ground is pretrained on O365v1, Table~\ref{tab:sota_coco} reports the pretraining data used by each COCO baseline for context, while Table~\ref{tab:sota_minilvis} focuses on miniLVIS results under the same O365v1 pretraining setting.

The standard OVOD results show that sentence-level grounding remains competitive beyond fine-grained evaluation. With O365v1 pretraining, SL-Ground achieves 48.8 AP on COCO and 25.27 AP on miniLVIS. Its COCO performance is close to the 49.5 AP reported by OV-DINO, while it achieves the highest overall and rare-category AP among the methods listed in Table \ref{tab:sota_minilvis}, reaching 25.27 AP and 18.52 AP$_r$, respectively. These results suggest that the sentence-level interface preserves general OVOD capability while improving fine-grained grounding.




\subsection{FG-OVD Benchmark}
\label{sub:fgovd}

Recent fine-grained OVOD studies have increasingly used the FG-OVD benchmark~\cite{bianchi2024devil} to evaluate whether open-vocabulary detectors can handle attribute-rich and long-form fine-grained descriptions beyond simple category names. Representative methods such as HA-FGOVD~\cite{ma2024hafgovd}, GUIDED~\cite{li2026guided}, and DSAA~\cite{jiang2026dsaa} improve fine-grained detection by introducing additional attribute-oriented modeling or adaptation strategies. We further examine the effectiveness of SL-Ground beyond our Stanford Cars setting on this common external benchmark.

DSAA~\cite{jiang2026dsaa} reports limited gains from directly fine-tuning Grounding DINO, and Table~\ref{tab:fgovd_comparison} shows its average AP increasing only from 33.5 to 34.0. Fine-tuning SL-Ground raises its average AP from 18.0 to 61.6, exceeding the reported DSAA results on all subsets except Transparency without adding FG-OVD-specific attribute-modeling modules. These results extend the evaluation to attribute-rich descriptions beyond Stanford Cars. Since the published baselines may differ in pretraining and implementation, this benchmark serves as an external performance comparison; the controlled variants in Sec.~\ref{sub:exp_stanfordcar} provide the more direct evidence for the grounding formulation.

\begin{table}[thp]
\centering
\caption{Comparison on FG-OVD benchmark.}
\label{tab:fgovd_comparison}
\resizebox{\linewidth}{!}{%
\begin{tabular}{lcccccccc|c}
\toprule
Detector & Hard & Med. & Easy & Triv. & Col. & Mat. & Pat. & Trans. & Avg. \\
\midrule
G-DINO & 17.0 & 28.4 & 31.0 & 62.5 & 41.4 & 30.3 & 31.0 & 26.2 & 33.5 \\
+ Finetune    & 16.6 & 28.0 & 24.3 & 73.6 & 42.7 & 29.4 & 29.3 & 28.4 & 34.0  \\
+ HA-FGOVD     & 19.2 & 32.3 & 34.0 & 62.2 & 41.5 & 33.0 & 32.1 & 29.2 & 35.4  \\
+ GUIDED       & 35.1 & 49.3 & 52.8 & 57.1 & 49.7 & 57.0 & 26.4 & 39.6 & 45.9  \\
+ DSAA         & 38.5 & 53.0 & 59.2 & 63.3 & 56.0 & 52.3 & 46.5 &  \textbf{63.0} & 54.0 \\
\midrule
SL-Ground      & 9.4 & 14.3 & 13.8 & 42.7 & 18.6 & 16.4 & 13.9 & 14.6 & 18.0 \\
+ Finetune     & \textbf{53.5} &  \textbf{64.4} &  \textbf{66.2} &  \textbf{71.8} &  \textbf{61.7} &  \textbf{66.3} &  \textbf{57.5} & 51.0 &  \textbf{61.6}  \\
\bottomrule
\end{tabular}
}
\par\smallskip
 {\footnotesize\raggedright
  Abbreviations: Med.=Medium, Triv.=Trivial, Col.=Color, Mat.=Material, Pat.=Pattern, and Trans.=Transparency.\par}
 
  \vspace{-5mm}
\end{table}

\subsection{Fine-Grained Recognition Diagnostic}
\label{sub:exp_finegrained}
We next complement the detection benchmarks with image-level fine-grained recognition on Stanford Dogs, CUB-200-2011, and Stanford Cars. This evaluation examines class discrimination among detections with sufficiently accurate localization and provides a complementary diagnostic of the sentence-level representation.

\subsubsection{Evaluation Protocol} The VLM baseline accuracies in Table~\ref{tab:finegrained_generalization} are reported by Finedefics~\cite{he2025analyzing}. Image-level top-1 accuracy is reported on Stanford Dogs (Dog-120)~\cite{khosla2011novel}, CUB-200-2011 (CUB-200)~\cite{wah2011caltech}, and Stanford Cars (Car-196), rather than detection AP.

For SL-Ground, we conduct a localization-controlled recognition diagnostic inspired by region-level recognition analyses such as RegionCLIP~\cite{zhong2022regionclip}. We keep final detections as candidates and retain only detections whose maximum IoU with any ground-truth box is at least 0.75. Each image-level class score is computed by class-wise max pooling over the retained detection scores~\cite{oquab2015object}. The highest-scoring class is used as the image-level prediction, while images with no retained detection are counted as incorrect. The resulting metric is image-level recognition accuracy under this localization criterion, rather than detection AP.

\subsubsection{Results and Interpretation}


\begin{table}[ht]

  \centering
\caption{Fine-grained recognition results on Dog-120, CUB-200, and Car-196. Values are accuracy (\%), not detection AP.}
  \label{tab:finegrained_generalization}
  \resizebox{\columnwidth}{!}{%
  \begin{tabular}{lccc}
    \toprule
    Model & Dog-120 & CUB-200 & Car-196 \\
    \midrule
    LLaVA 1.5~\cite{liu2024llava15} & 38.96 & 35.24 & 46.92 \\
    LLaVA-NeXT (Mistral)~\cite{liu2024llavanext} & 38.86 & 34.88 & 49.48 \\
    MobileVLM v2~\cite{chu2024mobilevlmv2} & 39.92 & 33.90 & 46.29 \\
    InstructBLIP Vicuna~\cite{dai2023instructblip} & 41.60 & 32.78 & 48.25 \\
    InstructBLIP Flan-T5-XL~\cite{dai2023instructblip} & 47.10 & 32.15 & 64.58 \\
    Phi-3-Vision~\cite{abdin2024phi3} & 39.80 & 37.63 & 54.50 \\
    BLIP2 Flan-T5-XL~\cite{li2023blip2} & 46.17 & 33.70 & 67.68 \\
    InternLM XComposer 2~\cite{dong2024internlmxcomposer2} & 41.47 & 37.42 & 53.89 \\
    Pali-Gemma~\cite{beyer2024paligemma} & 51.68 & 36.62 & 64.64 \\
    Idefics1~\cite{laurencon2023obelics} & 39.74 & 36.50 & 29.42 \\
    Idefics2~\cite{laurencon2024idefics2} & 57.96 & 47.17 & 80.25 \\
    Qwen-VL-Chat~\cite{bai2023qwenvl} & 66.18 & 52.30 & 76.23 \\
    Finedefics~\cite{he2025analyzing} & 72.86 & 57.61 & 84.67 \\
    \midrule
    SL-Ground$^\dagger$ & \textbf{86.54} & \textbf{84.74} & \textbf{85.09} \\
    \bottomrule
  \end{tabular}%
  }
  \par\smallskip
  {\footnotesize\raggedright
  $^\dagger$SL-Ground uses the procedure described above and has 182.4M parameters; the listed VLM baselines range from 3B to 10B parameters. \par}
  
\end{table}

SL-Ground achieves 86.54\%, 84.74\%, and 85.09\% accuracy on Dog-120, CUB-200, and Car-196, respectively. These results provide complementary evidence of fine-grained class discrimination under the stated localization criterion. The VLM results provide context, but the use of ground-truth boxes to select SL-Ground detections means that the table is not a comparison under an identical prediction pipeline. The detection results in Secs.~\ref{sub:exp_stanfordcar} and~\ref{sub:fgovd} provide the primary evidence for fine-grained OVOD performance.

\subsection{Effect of Training Candidate Set Size}
\label{sub:training_candidate_size}

To examine whether SL-Ground’s performance gain comes from using all negative categories during training, we apply the same 75-token candidate-sampling policy used by GD-CLIP to SL-Ground. GD-CLIP cannot fit all 189 category names into a single input sequence within its text encoder’s length limit, whereas SL-Ground encodes each category independently. We also evaluate fixed candidate-set sizes to assess the effect of the number of training candidates.

We define $K_{\mathrm{train}}$ as the number of candidate categories per image, including one ground-truth category and $K_{\mathrm{train}}-1$ negative categories. Under the 75-token budget, the candidate count varies with the tokenized lengths of the sampled category names. A one-pass simulation of this sampling policy over the 6,515 training images yields an average of 11.87 candidate categories per image. The fixed-size variants use $K_{\mathrm{train}}\in\{12,32,64,189\}$, with 189 corresponding to the complete vocabulary. All variants are evaluated using all 189 categories. All SL-Ground variants share the same O365v1 initialization, training configuration, random seed, and 20-epoch schedule.

\begin{table}[t]
\centering
\caption{Effect of training candidate-set size on Stanford Cars
car-model detection. Evaluation uses all 189 categories.}
\label{tab:training_candidate_size}
\small
\setlength{\tabcolsep}{3pt}
\renewcommand{\arraystretch}{1.08}
\begin{tabular*}{\columnwidth}{@{\extracolsep{\fill}}llcc@{}}
\toprule
Model & Training setting & $K_{\mathrm{train}}$ & AP $\uparrow$ \\
\midrule
GD-CLIP
    & 75-token sampling
    & 11.87$^{*}$
    & 14.5 \\
SL-Ground
    & 75-token sampling
    & 11.87$^{*}$
    & 70.6 \\
\midrule
SL-Ground
    & Fixed size
    & 12
    & 65.5 \\
SL-Ground
    & Fixed size
    & 32
    & 76.7 \\
SL-Ground
    & Fixed size
    & 64
    & 83.0 \\
SL-Ground
    & Full vocabulary
    & 189
    & \textbf{86.0} \\
\bottomrule
\end{tabular*}
\par\smallskip
{\footnotesize\raggedright
$^{*}$Mean candidate count from a one-pass sampling simulation over the 6,515 training images, including the ground-truth category; the count varies across images.
\par}
\end{table}

As shown in Table~\ref{tab:training_candidate_size}, SL-Ground achieves 70.6 AP under the same 75-token candidate-sampling policy used by GD-CLIP, compared with 14.5 AP for GD-CLIP. This gap indicates that SL-Ground's improvement cannot be explained solely by using more negative categories during training. This comparison matches training candidates only. At inference, GD-CLIP still processes the 189 descriptions in multiple prompt chunks, whereas SL-Ground scores all classes in a single pass. Full-vocabulary training further improves SL-Ground to 86.0 AP, indicating an additional benefit from using all candidate categories. Among the fixed-size variants, increasing $K_{\mathrm{train}}$ from 12 to 32, 64, and 189 improves AP from 65.5 to 76.7, 83.0, and 86.0, respectively, with diminishing gains. Using 64 candidates already brings SL-Ground within 3.0 AP of full-vocabulary training.

\subsection{Projected Embedding Space Analysis}
\label{sub:scoring_space}


To examine the representation differences underlying the Stanford Cars results, we analyze class separation in the projected embedding space. Restricting the analysis to queries matched to ground-truth objects reduces the influence of localization failures. For each matched query, we compute \(\Delta=d_{\mathrm{neg}}-d_{\mathrm{gt}}\), where \(d_{\mathrm{gt}}\) is the cosine distance to the ground-truth class text and \(d_{\mathrm{neg}}\) is the distance to the closest incorrect class text. A negative margin indicates a wrong-closer case, where the query is closer to an incorrect class text than to the ground-truth class text.

As shown in Fig.~\ref{fig:margin_cdf}, \(95\%\) and \(97\%\) of matched queries are wrong-closer for GD-BERT and GD-CLIP, against \(44\%\) for SL-Ground. These margins indicate better separation between ground-truth and competing class texts for SL-Ground in the analyzed projected space. Because the analysis is conditioned on matched object queries, it provides a representation diagnostic rather than a direct measure of detection accuracy or proof of the cause of all detection errors.

\color{black}



\subsection{Efficiency Analysis}
\label{sub:FPS}


Token-level grounding introduces extra inference cost because class names are serialized into long token sequences and split into multiple prompt chunks when they exceed the text encoder length limit. Each chunk requires a separate detector pass. In contrast, SL-Ground encodes each category as an independent sentence-level embedding and scores all candidate classes in one detector pass.

\begin{table}[t]
  \centering
 \caption{Accuracy--efficiency comparison on COCO and miniLVIS. Mem.\ denotes peak allocated GPU memory (GB).}
  \label{tab:efficiency}
  \resizebox{\columnwidth}{!}{%
  \begin{tabular}{lccccccc}
    \toprule
    \multirow{2}{*}{Model} &
    \multirow{2}{*}{Params (M) } &
    \multicolumn{3}{c}{COCO} &
    \multicolumn{3}{c}{miniLVIS} \\
    \cmidrule(lr){3-5} \cmidrule(lr){6-8}
    & & AP $\uparrow$ & FPS $\uparrow$ & Mem. $\downarrow$ 
      & AP $\uparrow$ & FPS $\uparrow$ & Mem. $\downarrow$ \\
    \midrule
    GD-BERT  & 173.0 & 48.4 & 12.38 & 1.37 & 15.78 & 0.58 & 1.34 \\
    GD-CLIP  & 187.1 & 45.6 & 6.28  & 1.27 & \textbf{26.37} & 0.26 & \textbf{1.25} \\
    SL-Ground  & 182.4 & \textbf{48.8} & \textbf{16.01} & \textbf{1.25} & 25.27 & \textbf{12.28} & 1.40 \\
    \bottomrule
  \end{tabular}%
  }
\end{table}


Table~\ref{tab:efficiency} shows that SL-Ground achieves the best AP, FPS, and peak memory on COCO. On miniLVIS, it trades 1.10 AP compared with GD-CLIP for a 47.2x FPS gain, and improves over GD-BERT by 9.49 AP and 21.2x FPS. Its peak memory is only slightly higher on miniLVIS, with at most 0.15 GB additional memory over the token-level variants. These results show that SL-Ground is particularly effective in large-vocabulary settings, where avoiding repeated token-level prompt chunks yields a large throughput gain.

\begin{figure}[ht]
  \centering
  \subfloat[Margin CDF\label{fig:margin_cdf}]{\includegraphics[width=0.495\linewidth]{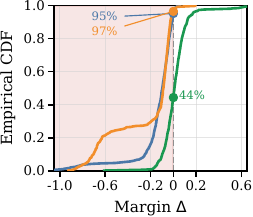}}%
  \subfloat[Latency scaling\label{fig:vocab_scaling_efficiency}]{\includegraphics[width=0.495\linewidth]{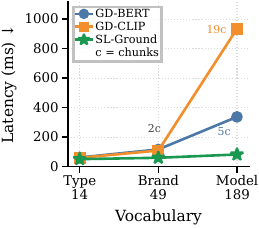}}
\caption{Stanford Cars margin CDF and vocabulary-scaling latency. Chunk labels denote the number of chunks in the tokenized prompt.}

  \label{fig:stanfordcar_margin_latency}
\end{figure}

Figure~\ref{fig:vocab_scaling_efficiency} further shows that the speed gap grows with vocabulary size. As labels extend from simple types to longer, more compositional descriptions like brand and car model, token-level variants require more prompt chunks and repeated detector passes, incurring a significant latency overhead. In contrast, SL-Ground keeps a single pass over sentence-level category embeddings, so its latency remains nearly flat across the same vocabulary growth.

\color{black}

\vspace{-3mm}
\vspace{3mm}

\subsubsection{Stage-Level Latency without Prompt Chunking}

\begin{figure}[t]
  \centering
  \subfloat[Stage composition\label{fig:stage_composition}]{\includegraphics[width=0.495\linewidth]{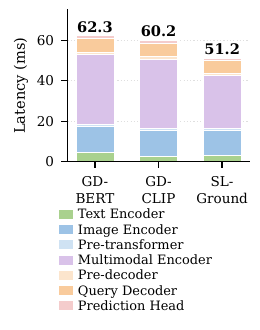}}%
  \hfill%
  \subfloat[Saving vs. GD-BERT\label{fig:module_saving}]{\includegraphics[width=0.495\linewidth]{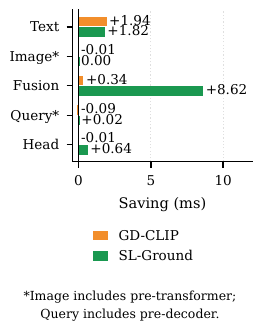}}
  \caption{Stage-level latency on Stanford Cars type labels, where all variants use one prompt chunk. Panel (a) shows stage composition, and panel (b) shows module-level latency savings relative to GD-BERT.}
  \label{fig:stage_delta_efficiency}
\end{figure}

To examine whether the speed advantage persists without repeated detector passes, Fig.~\ref{fig:stage_delta_efficiency} isolates the Stanford Cars type setting with 14 candidate classes, for which all three variants use a single prompt chunk. Figure~\ref{fig:stage_composition} reports the latency composition, with total latency of 62.3, 60.2, and 51.2 ms for GD-BERT, GD-CLIP, and SL-Ground, respectively. Figure~\ref{fig:module_saving} shows the savings relative to GD-BERT; multimodal fusion accounts for the largest reduction in SL-Ground. The token-level variants process a padded text sequence in the detector encoder, whereas SL-Ground uses 14 sentence-level class features. This reduces the text dimension involved in fusion, while the prediction head directly scores class features without token-to-class aggregation. The stage measurements therefore show that the efficiency advantage extends beyond avoiding prompt chunking.

\subsubsection{Memory--Latency Trade-off under Controlled Vocabulary Scaling}

\begin{figure}[t]
  \centering
  \subfloat[CUDA peak memory\label{fig:cuda_memory}]{\includegraphics[width=0.495\linewidth]{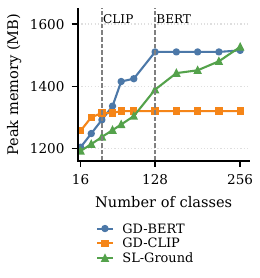}}%
  \hfill%
  \subfloat[Inference latency\label{fig:inference_latency}]{\includegraphics[width=0.495\linewidth]{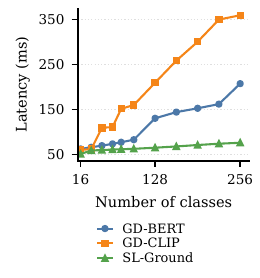}}
  \caption{Controlled memory--latency trade-off with synthetic single-token labels. Token-level variants cap per-pass CUDA memory after chunking but incur repeated detector passes, whereas SL-Ground keeps one pass with lower latency and class-count-dependent memory growth.}
  \label{fig:memory_latency_tradeoff}
\end{figure}

To separate the effect of vocabulary size from differences in category-description length, we evaluate CUDA peak memory and inference latency using synthetic class labels that each tokenize to a single token under both BERT and CLIP. Figure~\ref{fig:memory_latency_tradeoff} reports the resulting vocabulary sweep. Figure~\ref{fig:cuda_memory} shows lower peak memory for SL-Ground at small vocabulary sizes, followed by memory growth as more class features are processed together. The token-level variants limit the text sequence processed per pass through prompt chunking, but the additional detector passes increase latency, as shown in Fig.~\ref{fig:inference_latency}. SL-Ground instead scores all classes in one detector pass and maintains lower latency over the evaluated range. These results illustrate a memory--latency trade-off as the vocabulary grows, even when class-label token length is controlled.

\begin{figure*}[t]
  \centering
  {\small\bfseries Grounding DINO\par}
  \includegraphics[width=0.32\textwidth]{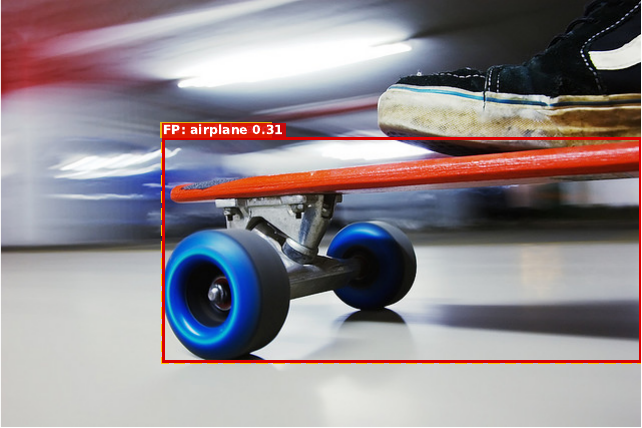}\hfill
  \includegraphics[width=0.32\textwidth]{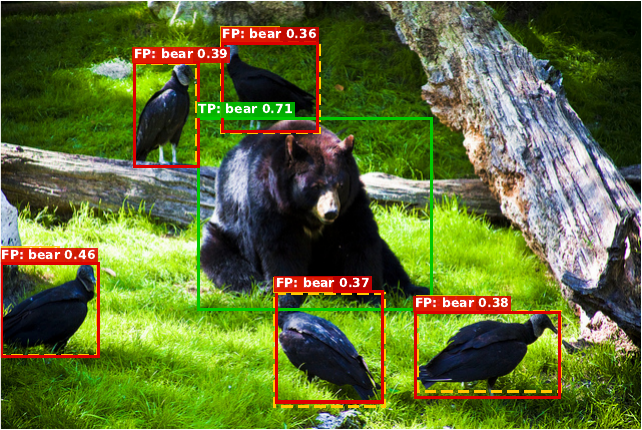}\hfill
  \includegraphics[width=0.32\textwidth]{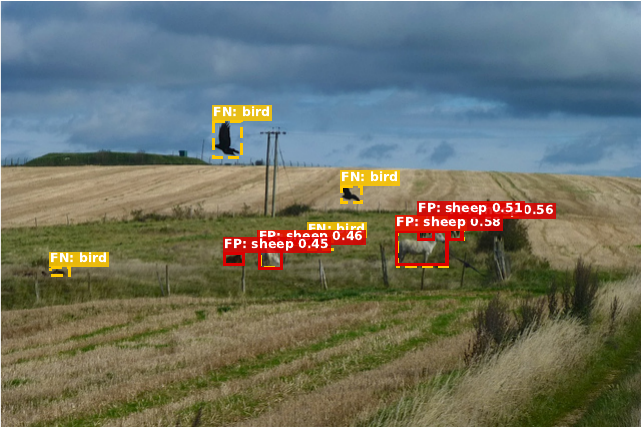}
  \vspace{1mm}
  {\small\bfseries SL-Ground\par}
  \includegraphics[width=0.32\textwidth]{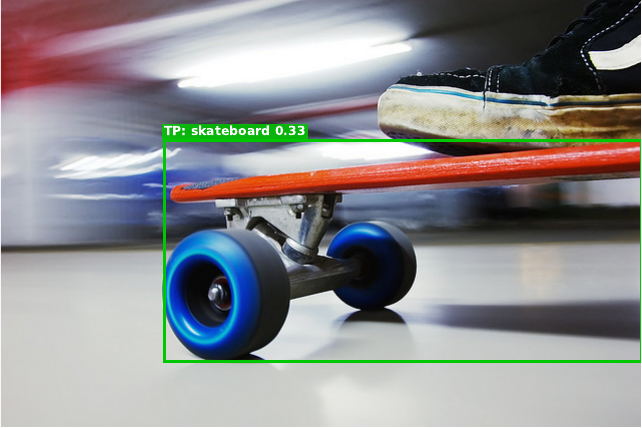}\hfill
  \includegraphics[width=0.32\textwidth]{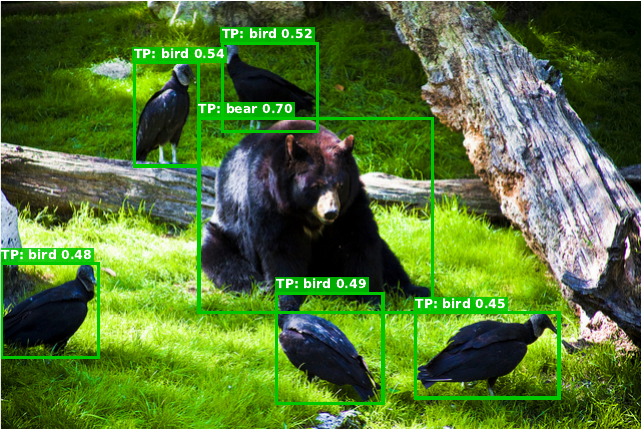}\hfill
  \includegraphics[width=0.32\textwidth]{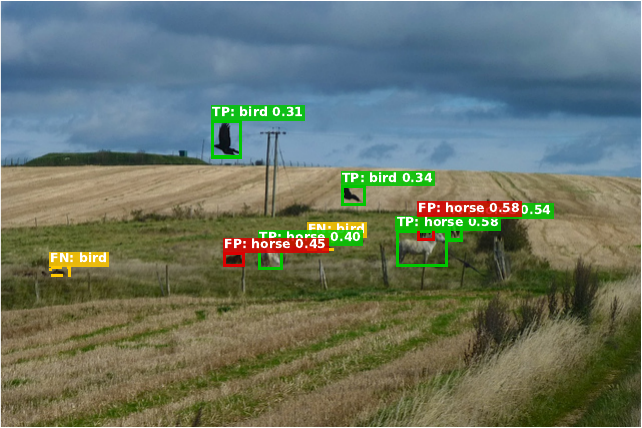}
  \caption{Qualitative comparison on COCO. Grounding DINO is shown in the top row and SL-Ground in the bottom row. SL-Ground reduces class confusion in examples such as skateboard, bird, and horse while preserving comparable box locations.}
  \label{fig:supp_coco_qualitative}
\end{figure*}

\subsection{Qualitative Analysis}
\label{sub:qualitative}

\begin{figure*}[!t]
  \centering
  {\small\bfseries Grounding DINO\par}
  \includegraphics[width=0.35\textwidth]{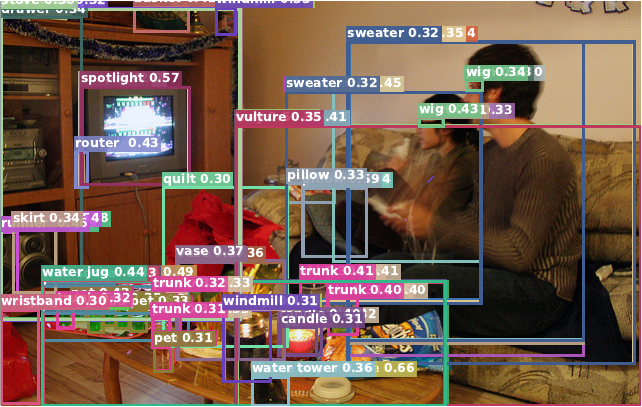}\hfill
  \includegraphics[width=0.295\textwidth]{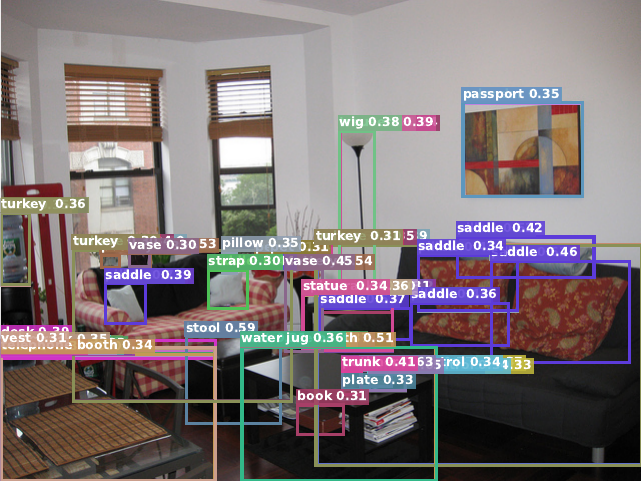}\hfill
  \includegraphics[width=0.334\textwidth]{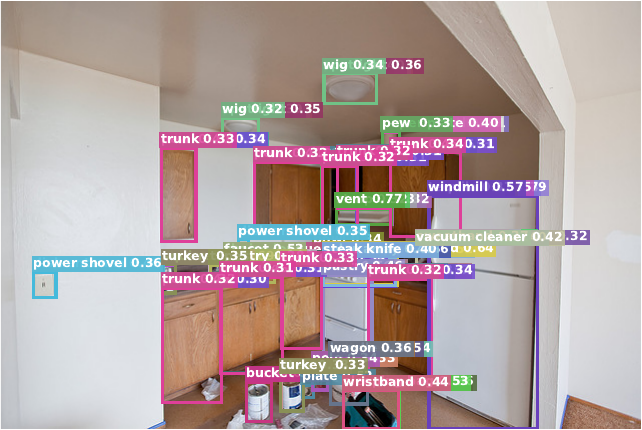}
  \vspace{1mm}
  {\small\bfseries SL-Ground\par}
  \includegraphics[width=0.35\textwidth]{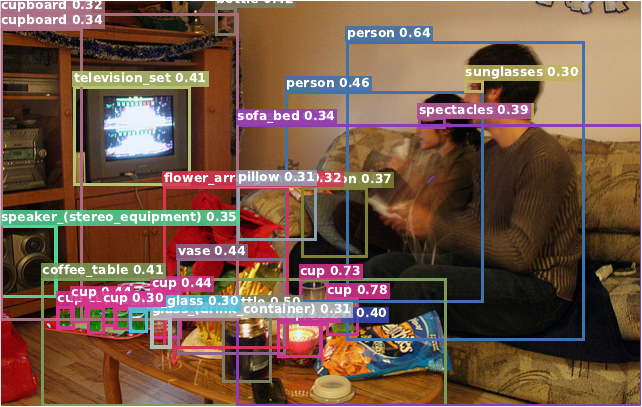}\hfill
  \includegraphics[width=0.295\textwidth]{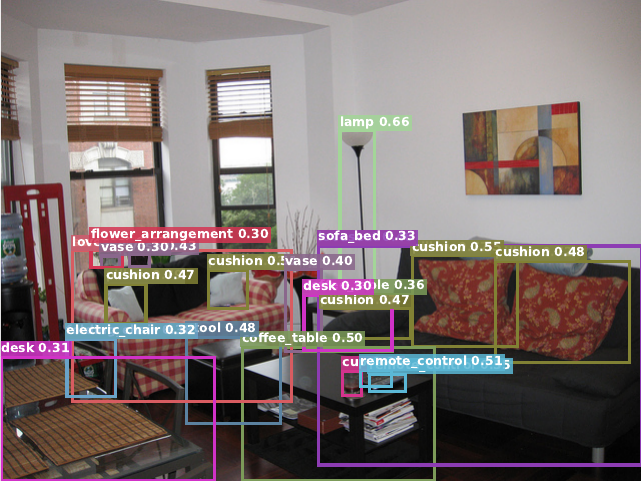}\hfill
  \includegraphics[width=0.334\textwidth]{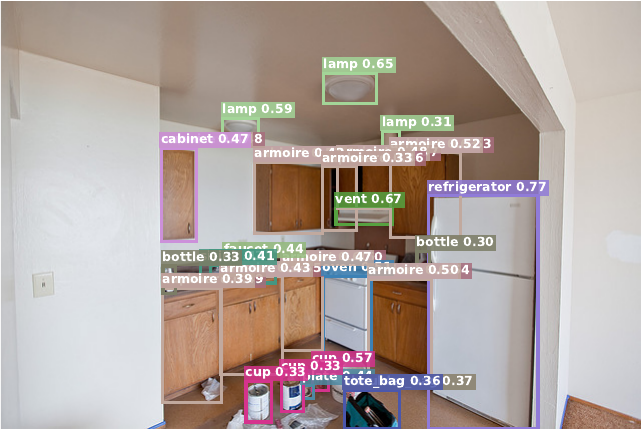}
 \caption{Qualitative comparisons on miniLVIS with 1,203 candidate categories. Grounding DINO is shown in the top row and SL-Ground in the bottom row. Grounding DINO shows overlapping predictions with different class labels after outputs from multiple prompt chunks are concatenated without cross-chunk suppression, while SL-Ground produces a less dense set of predictions in a single pass.}

  \label{fig:supp_minilvis_qualitative_additional}
\end{figure*}

Finally, we visualize predictions to examine class confusion and the effects of prompt chunking. \Cref{fig:supp_coco_qualitative,fig:supp_minilvis_qualitative_additional} compare Grounding DINO (GD-BERT)~\cite{liu2024grounding} and SL-Ground on matched COCO and miniLVIS images.

\subsubsection{Qualitative Protocol}
COCO~\cite{lin2014microsoft} and miniLVIS~\cite{gupta2019lvis} use 80 and 1,203 candidate categories, respectively. For both methods, each prediction-head invocation keeps at most 300 predictions via \texttt{max\_per\_img}; the final number of candidates differs when outputs from multiple prompt chunks are combined, as described below. A confidence threshold of 0.30 is applied only when visualizing predictions. In the COCO examples, green, red, and yellow boxes denote true positives, false positives, and false negatives, respectively.

\subsubsection{Chunking Behavior}

Grounding DINO serializes category descriptions into prompt chunks when they exceed the text encoder limit; miniLVIS requires 31 chunks. Each chunk uses a separate detector pass and independently retains its top-300 predictions. The chunk outputs are concatenated without global top-k selection, cross-chunk non-maximum suppression, score filtering, or duplicate removal, producing up to 9,300 predictions before visualization. In contrast, SL-Ground encodes each candidate category independently, scores all classes in one detector pass, and applies \texttt{max\_per\_img} once, producing at most 300 predictions before visualization.

\subsubsection{Qualitative Observations}
\Cref{fig:supp_coco_qualitative} illustrates examples of reduced class confusion under a moderate vocabulary, while \Cref{fig:supp_minilvis_qualitative_additional} shows the denser predictions produced by repeated prompt chunking under a larger vocabulary. SL-Ground avoids cross-chunk concatenation by scoring all candidate classes in one pass. The difference in prediction density partly reflects this output-processing difference, so fewer displayed boxes alone do not establish better recognition or localization. These examples complement the quantitative detection results rather than demonstrate a universal localization advantage.

\raggedbottom
\section{Conclusion and Future Work}



This work studies how sentence-level text representations can affect grounding-based OVOD for fine-grained category prediction. The controlled comparison shows that replacing the text encoder alone yields limited gains at the finest granularity, whereas adopting sentence-level features throughout the grounding pipeline substantially improves performance. SL-Ground addresses this bottleneck by keeping each category description as a sentence-level representation and scoring object queries directly against complete category descriptions. 
The improvement persists under matched training candidate sampling, while using the full vocabulary provides an additional benefit. SL-Ground also better preserves general open-vocabulary performance after fine-grained adaptation and avoids repeated prompt chunking as the vocabulary grows. These results support the importance of the text representation throughout fine-grained grounding-based OVOD, especially when categories are compositional and share tokens. Future work will further characterize where sentence-level grounding is most beneficial, and where tasks that require word-level correspondence, such as referring expressions or attribute localization, require token-level or hybrid designs.

\section*{Acknowledgment}
This work was supported by the National Science and Technology Council (NSTC), Taiwan, under grant number NSTC 114-2221-E-194-030-MY3. We thank the National Center for High-performance Computing (NCHC), NIAR for providing computational and storage resources.
\FloatBarrier
\bibliographystyle{IEEEtran}

\bibliography{main}

\vspace{-1.5em}

\begin{IEEEbiography}[{\includegraphics[width=1in,height=1.25in, trim=170 500 150 150,clip,keepaspectratio]{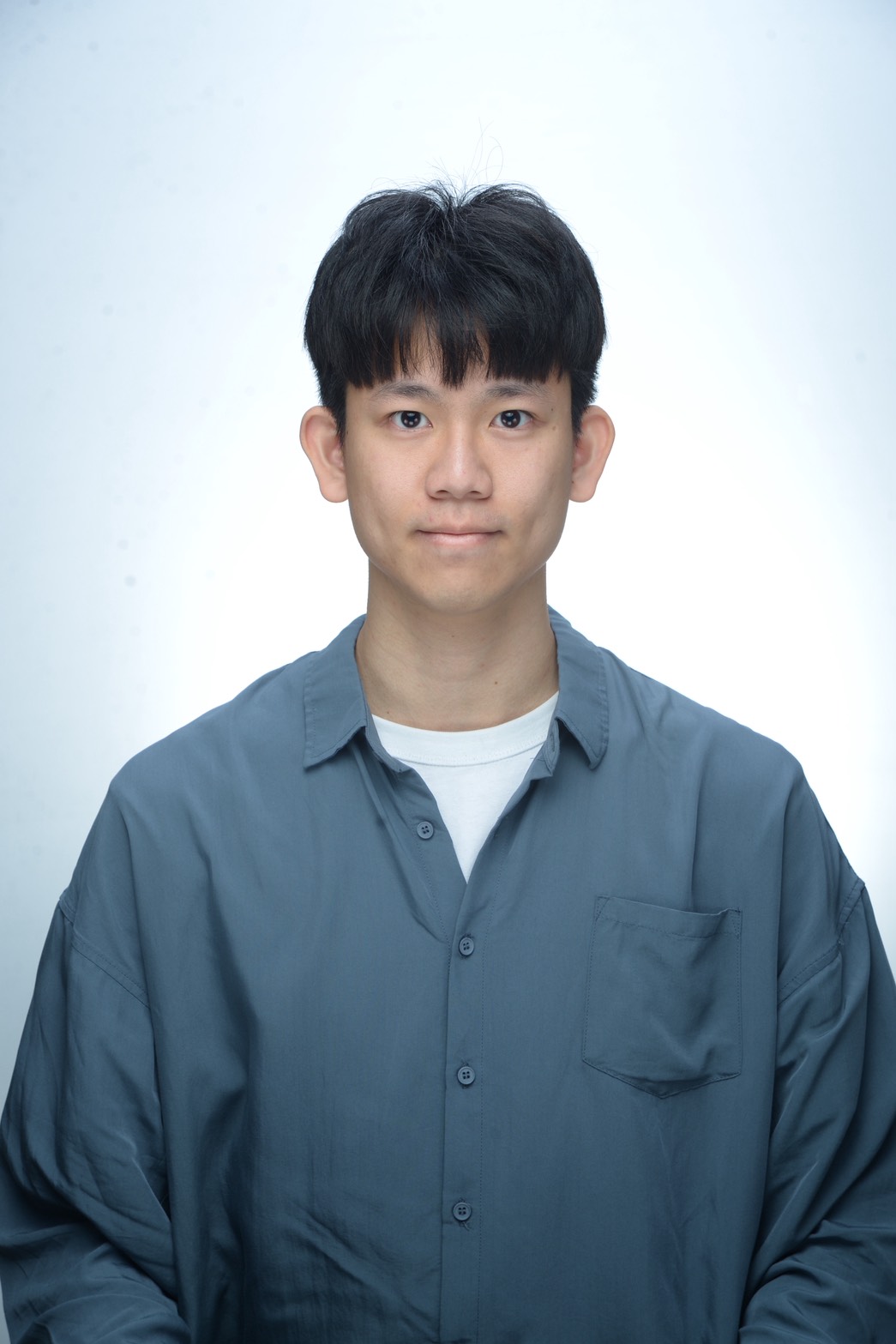}}]{Yang-Syu Li}
received the M.S. degree in computer science and information engineering from National Chung Cheng University, Chiayi, Taiwan. His research interests include computer vision, deep learning, and object detection.
\mbox{}
\end{IEEEbiography}

\vspace{-1.5em}

\begin{IEEEbiography}[{\includegraphics[width=1in,height=1.25in,clip,keepaspectratio]{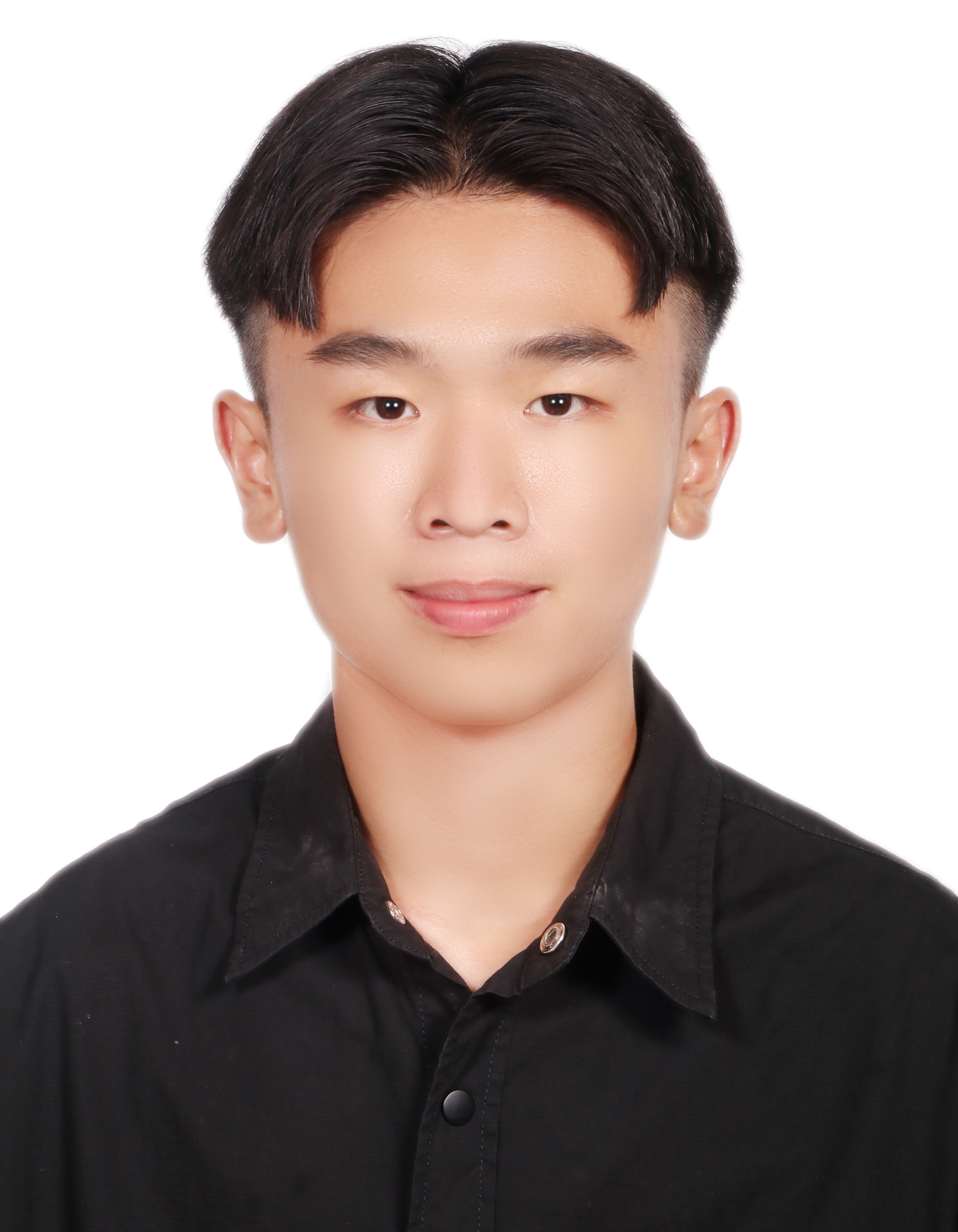}}]{Cheng-Huan Yang}
is currently pursuing the B.S. degree in computer science and information engineering at National Chung Cheng University, Chiayi, Taiwan. His research interests include computer vision, multimodal learning, and machine learning.
\end{IEEEbiography}

\vspace{-1.5em}

\begin{IEEEbiography}[{\includegraphics[width=1in,height=1.25in,clip,keepaspectratio]{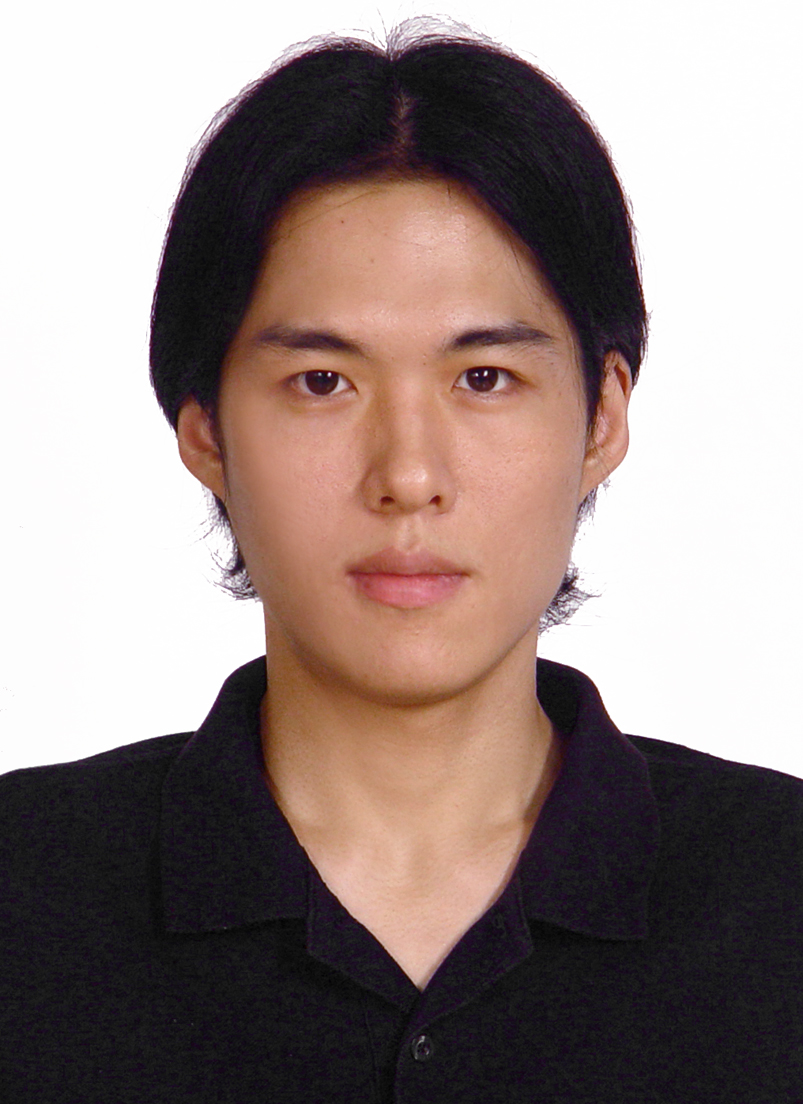}}]{Chia-Yung Jui}
received his B.S. and M.S. degrees in Department of Applied Mathematics in National Chung Hsing University. He is currently working at the National Center for High-Performance Computing. His research interests include applied numerical methods, deep learning and interdisciplinary applications of numerical and machine learning methods.
\end{IEEEbiography}

\vspace{-1.5em}

\begin{IEEEbiography}[{\includegraphics[width=1in,height=1.25in,clip,keepaspectratio]{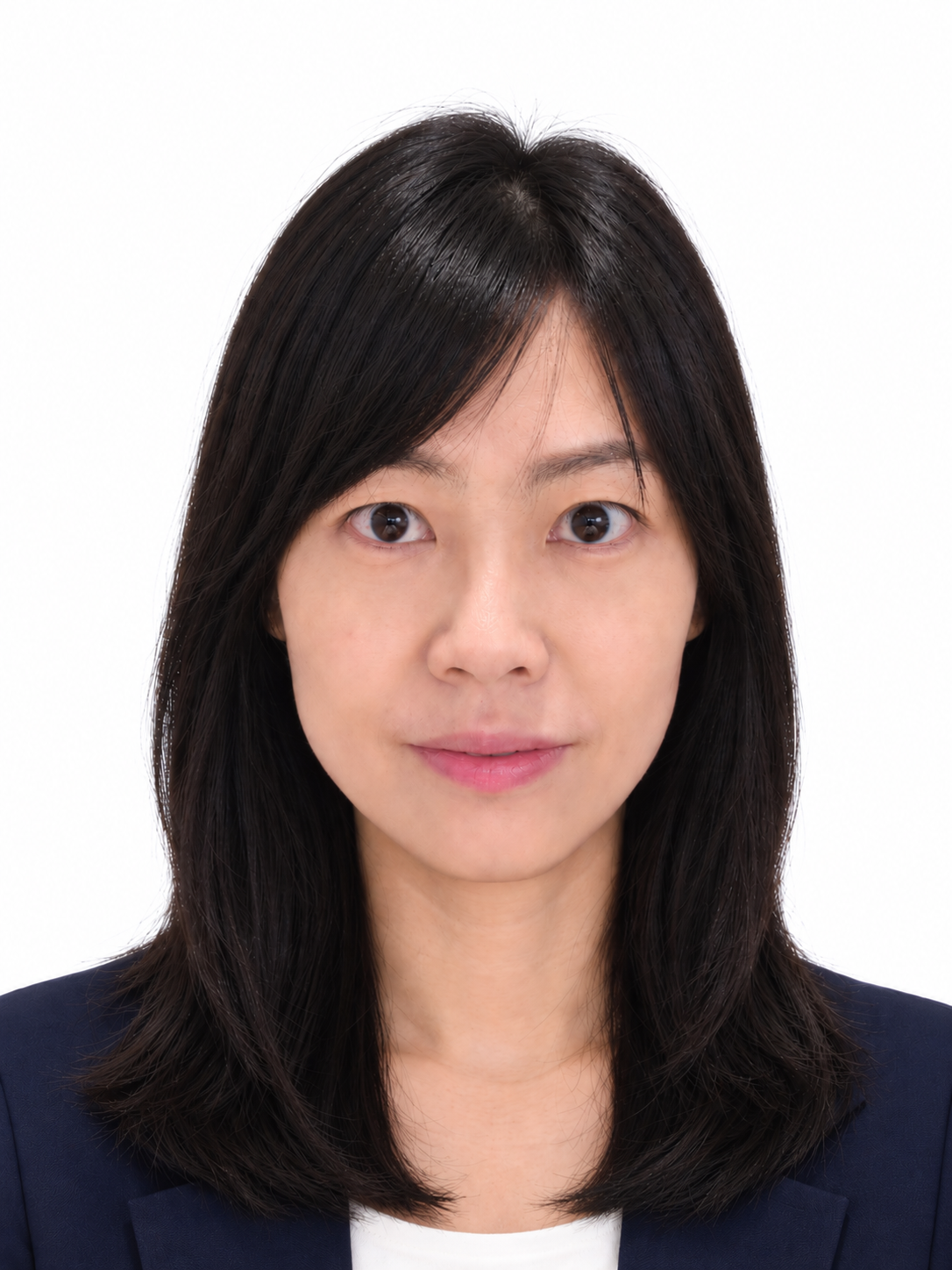}}]{Peggy Joy Lu}
received her Ph.D in Computer Science at National Yang Ming Chiao Tung University, Taiwan in 2023, the B.S. and M.S. degrees, both in Computer Science and Information Engineering Department in National Chiao Tung University in 2009 and 2011, respectively. She is currently an assistant professor at the National Chung Cheng University, Taiwan. Her research interests include computer vision, multimodal learning, visual recognition, domain adaptation and generalization, and vision-language models.
\end{IEEEbiography}

\end{document}